\documentclass[conference]{IEEEtran}
\IEEEoverridecommandlockouts
\usepackage{cite}
\usepackage{amsmath,amssymb,amsfonts}
\usepackage{algorithmic}
\usepackage{graphicx}
\usepackage{textcomp}
\usepackage{xcolor}
\usepackage{booktabs}
\usepackage{hyperref}
\usepackage{xspace}

\usepackage[linesnumbered,ruled,vlined]{algorithm2e}

\def\BibTeX{{\rm B\kern-.05em{\sc i\kern-.025em b}\kern-.08em
    T\kern-.1667em\lower.7ex\hbox{E}\kern-.125emX}}
\begin{document}

\title{Poster: Verifiable Unlearning on Edge}

\author{\IEEEauthorblockN{Mohammad M Maheri}
\IEEEauthorblockA{\textit{Imperial College London} \\
}
\and
\IEEEauthorblockN{Alex Davidson}
\IEEEauthorblockA{\textit{LASIGE, Faculdade de Ciências, Universidade de Lisboa} \\
}
\and
\IEEEauthorblockN{Hamed Haddadi}
\IEEEauthorblockA{\textit{Imperial College London} \\
}
}




\newcommand{\forgetset}{\textit{forget set}\xspace}
\newcommand{\forgetclass}{\textit{forget class}\xspace}
\newcommand{\retainset}{\textit{retain set}\xspace}

\maketitle

\begin{abstract}
Machine learning providers commonly distribute global models to edge devices, which subsequently personalize these models using local data. However, issues such as copyright infringements, biases, or regulatory requirements may require the verifiable removal of certain data samples across all edge devices. Ensuring that edge devices correctly execute such unlearning operations is critical to maintaining integrity.

In this work, we introduce a verification framework leveraging zero-knowledge proofs, specifically zk-SNARKs, to confirm data unlearning on personalized edge-device models without compromising privacy. We have developed algorithms explicitly designed to facilitate unlearning operations that are compatible with efficient zk-SNARK proof generation, ensuring minimal computational and memory overhead suitable for constrained edge environments. Furthermore, our approach carefully preserves personalized enhancements on edge devices, maintaining model performance post-unlearning.

Our results affirm the practicality and effectiveness of this verification framework, demonstrating verifiable unlearning with minimal degradation in personalization-induced performance improvements. Our methodology ensures verifiable, privacy-preserving, and effective machine unlearning across edge devices.



\end{abstract}


\section{Introduction}

 Machine unlearning aims to erase the influence of specific data points from trained models, addressing privacy regulations such as GDPR, which grant individuals the “right to be forgotten.” 
 Ensuring that unlearning has been done correctly, however, is challenging. The client, who holds a personalized model, must prove to the model provider that the requested data has been removed—without revealing any sensitive information to the model provider. This is especially difficult if the client might be dishonest or unwilling to perform the unlearning faithfully.
 This challenge is exacerbated on edge devices, where users personalize centralized models locally with sensitive private data that should neither be disclosed nor transferred elsewhere. Consequently, proving that a data point has been successfully unlearnt on such personalized models demands privacy-preserving verification mechanisms. Recently, zero-knowledge succinct non-interactive arguments of knowledge (zk-SNARKs) have emerged as a promising cryptographic technique for verifying computations, particularly in deep neural network inference~\cite{maheri2025telesparse}. zk-SNARKs thus represent a potential solution by enabling edge devices (acting as provers) to demonstrate the correctness of their local unlearning computations without revealing underlying private data or personalized model parameters.

 However, personalized model unlearning on edge devices introduces unique challenges. Unlike centralized settings where the model provider can compute and distribute unlearning updates, applying such updates directly to personalized models can severely degrade their tailored performance, undermining the benefits of personalization. Furthermore, naively sending the raw unlearning data (i.e., the \forgetset) to the client and requesting proof of unlearning is infeasible: i) it exposes the \forgetset, compromising its privacy, and ii) it requires clients to generate zkSNARK proofs for full unlearning algorithms—such as multi-epoch gradient ascent or retraining on the retained dataset—which demands prohibitively high computation and memory, making it impractical for edge devices. Generating zkSNARK proofs for such heavy procedures is particularly unrealistic, as proof generation for even moderate-scale training remains orders of magnitude more expensive than inference-level proofs~\cite{sun2024zkdl}. These limitations highlight the need for a new approach that enables efficient and privacy-preserving verification of unlearning on locally personalized models, without revealing sensitive data or compromising model utility. In this work, we address this gap by designing a zkSNARK-friendly approximate unlearning procedure tailored for personalized models on edge devices.


\section{Methodology}

\begin{figure}
    \centering
    \includegraphics[width=0.90\linewidth]{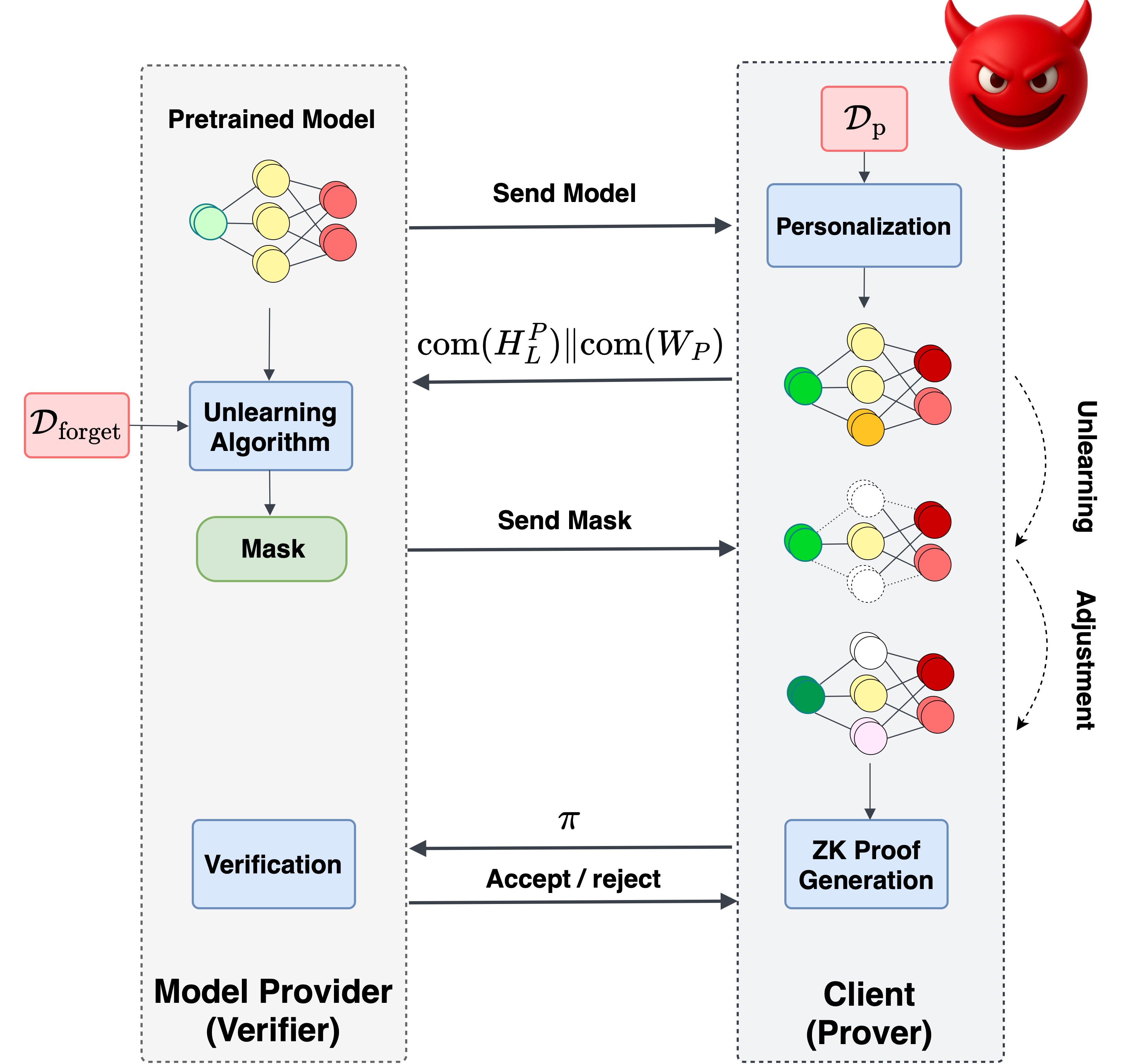}
    \caption{Framework overview of the proposed privacy-preserving method.}
    \label{fig:protocol}
\end{figure}

Our methodology—outlined in Figure~\ref{fig:protocol}—proceeds in two tightly integrated stages.
(i) Approximate machine unlearning is achieved by pruning neurons predominantly activated by the \forgetset, followed by Optimal Brain Surgeon (OBS)-based weight adjustment tailored to the user’s personalized loss landscape.
(ii) A zkSNARK proof is then generated for the proposed unlearning procedure, which combines pruning and weight adjustment. Since the procedure is designed to be efficient and zk-friendly, it enables practical proof generation on resource-constrained clients.

\subsection{Unlearning Algorithm}

Selective pruning provides a promising avenue for approximate machine unlearning~\cite{pochinkov2024dissecting}, as it identifies and removes model components (e.g., neurons or attention heads) that are disproportionately influential on the targeted \forgetset, while preserving performance on the \retainset.
Formally, let $D_{\text{forget}}$ and $D_{\text{retain}}$ denote the forget and retain datasets, respectively. For each neuron $n$, an importance score is computed as
\[
\text{Score}(n; D_{\text{retain}}, D_{\text{forget}}) = \frac{\text{Importance}(D_{\text{forget}}, n)}{\text{Importance}(D_{\text{retain}}, n) + \epsilon},
\]
where $\text{Importance}(D, n)$ quantifies the average magnitude of neuron $n$'s activations over dataset $D$. Neurons with the highest scores are pruned by zeroing their associated parameters, thereby selectively removing the influence of the \forgetset.



While pruning yields a binary mask that can be applied to remove parameters and induce forgetting, directly applying such a mask to a personalized model is problematic. Personalized models have adapted their parameters to sensitive local data; thus, indiscriminate pruning based on global importance scores can erase features critical for the personalization task, resulting in significant accuracy degradation. This phenomenon arises because pruning typically targets neurons influential to the \forgetset, without considering their contribution to the retained personalized data, as neural representations often entangle multiple tasks and data sources.

To mitigate this degradation, we leverage the OBS framework~\cite{lecun1989optimal,hassibi1993optimal,kuznedelev2023cap}, which uses second-order information to optimally adjust the remaining weights after pruning. Formally, starting from a dense parameter vector $w^*$, the post-pruning parameters $w_M$ are determined by minimizing the second-order Taylor approximation of the loss:
\[
L(w_M) - L(w^*) \approx \frac{1}{2}(w_M - w^*)^\top H_L(w^*)(w_M - w^*),
\]
where $H_L(w^*)$ is the Hessian matrix of the loss evaluated at $w^*$.  
Therefore, for a pruned weight $i$, the optimal adjustment to the remaining weights is given by:
\begin{equation}
    \label{eq:weight_compensation}
    \delta w^* = - \frac{w_i}{[H_L(w^*)^{-1}]_{ii}} H_L(w^*)^{-1} e_i,
\end{equation}
where $e_i$ is the unit vector corresponding to weight $i$.

Building on this principle, some recent works ~\cite{kurtic2022optimal,kuznedelev2023cap} introduce block-wise pruning for computational efficiency. In our adaptation, we compute the empirical Fisher information matrix on the personalized dataset to accurately capture the local curvature relevant to the user's data. This ensures that pruning-induced perturbations are adjusted to preserve performance on the personalization dataset, while still achieving approximate unlearning of the targeted \forgetset.

\subsection{Efficient Zero-Knowledge Proof Generation}


\begin{algorithm}[t]
\small
\DontPrintSemicolon
\SetAlgoLined
\SetNlSty{}{}{} 
\caption{Verifiable Approximate Unlearning}
\label{alg:verifiable_unlearning}
\KwIn{Public parameters $\textsf{pub}$, initial personalized model commitment $\mathsf{com}_P$, pruning mask $\mathsf{mask}$, Fisher matrix commitment $\mathsf{com}_H$}
\KwOut{New model commitment $\mathsf{com}_P'$, zkSNARK proof $\pi$}
\vspace{0.8em}
\textbf{Client Setup:} \\
Commit to Fisher matrix $H_L^P$, obtain $\mathsf{com}_H$  {\scriptsize
(Offline step)}

\vspace{0.5em}
\textbf{Unlearning Request:} \\
Receive pruning mask $\mathsf{mask}$ from model provider

\vspace{0.5em}
\textbf{Model Weight Adjustment:} \\
\For{each pruned weight $i$ in $\mathsf{mask}$}{
    Compute weight adjustment $\delta w_i^*$ using Equation~\eqref{eq:weight_compensation}
}

\vspace{0.5em}
\textbf{New Model Computation:} \\
Update model weights: $w_P' = w_P + \delta w^*$

\vspace{0.5em}
\textbf{Proof Generation:} \\
Generate zkSNARK proof $\pi$ that:\\
\quad (i) $\delta w^*$ computed correctly from $\mathsf{com}_H$ and $\mathsf{mask}$\\
\quad (ii) $\mathsf{com}_P' = \mathsf{com}_P + \delta w^*$

\vspace{0.5em}
\textbf{Proof Verification:} \\
Verifier checks $\pi$ and updated commitment $\mathsf{com}_P'$
\end{algorithm}

Building on the pruning-based approximate unlearning strategy, we now describe how the proposed method enables efficient zkSNARK proof generation on edge devices. Instead of requiring the prover (i.e., the client) to generate a proof for the full fine-tuning or retraining process—which would be computationally prohibitive—we design a lightweight proof structure tailored to the unlearning procedure. Each client commits to the empirical Fisher information matrix $H_L^P$ computed on their personalized dataset. Upon receiving the unlearning request (specified as a pruning mask), the prover needs only to demonstrate the model update according to the weight adjustment in Equation~\eqref{eq:weight_compensation}. The overall interaction between prover and verifier is illustrated in Figure~\ref{fig:protocol} and Algorithm~\ref{alg:verifiable_unlearning}.

Since weight updates are performed block--wise ----with $H_L^P$ structured as a block diagonal matrix ---- the proof generation decomposes into verifying independent matrix vector multiplications within each block. This structure is highly zk-friendly, as it avoids non-linear operations that are difficult to verify and allows block-wise proofs to be generated sequentially or in parallel. After applying the weight adjustment updates, the client proves that the new model commitment corresponds to the sum of the previous model commitment and the sparse model update. Crucially, the actual model parameters and the specific updates remain completely hidden from the model provider (the verifier), ensuring that privacy is preserved throughout the verification process of unlearning.

\begin{table*}[!ht]
\centering
\caption{Impact of Unlearning and Weight Adjustment on Personalized Model Performance}
\begin{tabular}{lcc}
\toprule
\textbf{Method} & \textbf{Forget Class Accuracy (\%) $\downarrow$} & \textbf{Personalized Accuracy (\%) $\uparrow$} \\
\midrule
Personalized Model (Baseline) & 93.7 & 71.5 \\
After Applying Unlearning Mask & 60.5 & 69.4 \\
After Weight Adjustment (\textbf{Our method}) & 59.5 & 70.9 \\
\midrule
\textit{Improvement over Naive Mask (\%)} & \textbf{1.7} & \textbf{71.4} \\
\bottomrule
\end{tabular}
\label{table:performance}
\end{table*}

\section{Evaluation}


Our experimental evaluation aims to assess the effectiveness of the proposed unlearning algorithm in personalized settings. Specifically, we focus on: (i) evaluating the extent to which the unlearning algorithm removes information pertaining to the \forgetclass on the personalized model by measuring the model's accuracy on that class; (ii) quantifying the impact of directly applying the unlearning mask on the personalized model's performance; and (iii) determining the degree to which the weight adjustment compensates for any performance degradation resulting from the unlearning process.


To simulate this scenario, we utilize a Vision Transformer (ViT) model pretrained on the ImageNet dataset. The model provider selects the "birds" class as the \forgetset and computes the corresponding unlearning mask based on the pretrained model. 
The mask modifies only the Multi-Layer Perceptron (MLP) sublayers within the transformer blocks, pruning 2\% of their parameters while leaving all other model parameters unchanged.
To emulate personalization, each client fine-tunes the model on a subset of the ImageNet-Sketch dataset~\cite{wang2019learning}, focusing on specific classes such as "fish."
ImageNet-Sketch introduces a domain shift through sketch-style images, challenging the model to generalize across visual modalities while preserving class semantics---making it a strong benchmark for studying personalization under distributional variation.
Personalization is performed by fine-tuning the upper layers of the ViT model, using a layer-wise learning rate decay while freezing early transformer blocks. This allows the model to adapt to client-specific data without destabilizing general representations.




In our experiments, we evaluate the effectiveness of the proposed unlearning procedure in removing the influence of the \forgetset while preserving personalized model utility. Table~\ref{table:performance} shows the personalized model’s performance before unlearning, after applying the pruning-based mask, and after our OBS-based weight adjustment.

Applying the unlearning mask alone substantially reduces the model's accuracy on the \forgetclass (from 93.7\% to 60.5\%), indicating that the pruning successfully disrupts the model’s reliance on the targeted data. However, this step also introduces a non-negligible performance drop on the client’s personalized data (from 71.5\% to 69.4\%), underscoring the trade-off between forgetting and utility preservation.

Our weight adjustment method restores most of the lost personalized accuracy (up to 70.9\%) and further lowers the \forgetclass accuracy to 59.5\%. Measured relative to the masked model, these gains highlight our method’s ability to enhance forgetting while recovering utility. Notably, this is achieved by modifying only 2\% of the MLP parameters.

\section{Conclusion and Future Work}

We proposed an initial framework to approach the challenge of verifiable machine unlearning for personalized models on edge devices, motivated by the need for privacy-preserving compliance with unlearning requests. We proposed an efficient pruning-based approximate unlearning method based on OBS adjustment, tailored for zkSNARK-friendly proof generation.


Experiments on personalized ViT models fine-tuned on ImageNet-Sketch demonstrate that the proposed method enables effective unlearning, recovering over 70\% of the personalized accuracy lost due to naive pruning, while maintaining minimal degradation in downstream utility.

Future work includes a concrete evaluation of proof generation costs—covering computation time and memory usage on edge devices—compared against baselines such as naive retraining verification or approximate unlearning methods not designed for zk-friendliness. Additionally, extending our method to large-scale personalized models such as large language models (LLMs), where block-diagonal Fisher approximation and sparse updates are critical, is a promising direction to further validate scalability and effectiveness.



\section*{Acknowledgment}
This work was supported by the UKRI Open Plus Fellowship (EP/W005271/1), the EU CHIST-ERA GRAPHS4SEC project, and FCT through the ParSec project (2024.07643.IACDC) and LASIGE (UID/00408/2025).



\bibliographystyle{ieeetr} 
\bibliography{references}  

\end{document}